\documentclass[]{spie}  

\usepackage{amsmath,amsfonts,amssymb}
\usepackage{graphicx}
\usepackage[colorlinks=true, allcolors=blue]{hyperref}
\usepackage{caption}
\usepackage{float}
\usepackage{todonotes}

\begin{document}

\title{Giving AI a Headache: Acoustic Adversarial Attacks to Computer Vision Applications}

\author[a]{Nicole Villavicencio-Garduño}
\author[b]{Maksim Ekin Eren}
\author[c]{Milo Prisbrey}
\author[d]{Ben Migliori}
\author[e]{Michael Teti}
\affil[a]{Carnegie Mellon University, Pittsburgh, USA}
\affil[a, b, c, d, e]{Los Alamos National Laboratory, Los Alamos, USA}

\authorinfo{Further author information: (Send correspondence to M.T.)\\M.T.: E-mail: mteti[at]lanl[dot]gov}

\pagestyle{empty} 
\setcounter{page}{301} 
 
\maketitle

\begin{abstract}
Artificial Intelligence (AI) is increasingly used to automate a variety of real-world computer vision (CV) applications, such as autonomous vehicle control, facial recognition, and security cameras. Recent research has shown that acoustic vibration can induce real physical motion in cameras, interfering with their internal stabilization mechanisms. Because the motion falls outside the conditions the stabilization system was designed to handle, the system introduces artifacts into the frame, causing AI-based CV models to misclassify, miss targets, or hallucinate objects.

Previous work used ultrasonic frequencies ($>$20 kHz) to perform short-range attacks, which limits them to short distances due to the attenuation exhibited by high frequencies. In this work, we investigate acoustic attacks using lower frequencies in the audible range ($<$20 kHz), and we further expand our analysis to include how various image and object features are affected by the attacks. 

Specifically, we performed physical experiments to demonstrate the viability of our attacks on an off-the-shelf object detection model (YOLO11) by resonating a commercially available camera with various frequencies. Based on our results, we provide insights into several factors that make an AI CV system more vulnerable to these attacks, which could help inform the development of future mitigation strategies.
\end{abstract}


\section{INTRODUCTION}
\label{sec:intro}  

Computer Vision (CV) is a mature domain of artificial intelligence (AI) that  extracts information from images, videos, and other visual inputs. This enables systems to make decisions based on visual data, often in critical applications. As these systems are increasingly deployed in real-world settings and may operate without direct human oversight, ensuring their accuracy and reliability is essential to providing safe and reliable systems.

Recently, researchers have examined the effects of acoustic excitation on cameras backed by vision-based decision making systems. They showed that the oscillations induced by the acoustic pressure  can heavily impact imaging performance, which can introduce complex challenges to image stabilization methods. 

The role of image stabilization sensors is to drive movement of either imaging sensors or conjugate plane lens elements to counteract motion of the camera frame. Typically, these systems are designed to correct for small, low frequency system disturbances introduced by operator movement or other sources.  By targeting stabilization systems with vibrations they cannot correct for (and are not designed to account for), adversarial artifacts in images can be created via acoustic means and cause distortions that deceive an object detection model.

Previous research findings fail to focus on novel acoustic attacks on camera sensors \cite{Ji2021PoltergeistAA}. There have been no attempts by camera system developers to examine larger plausible frequency ranges such as those found in these attacks. Previous research also fails to explore lower frequency ranges, which is important because lower frequencies having longer wavelengths, allowing them to diffract more easily around objects. Low frequencies may travel further as well, depending on the specific environment. This makes them especially relevant in real-world environments, where sound waves encounter obstructions.

In this paper, we expand the idea of acoustic attacks across a broader range of frequencies, environments, and object detection models then previously studied in the literature. We use a YOLOv11-based object detection system and an off-the-shelf camera capturing live footage to approximate real-world conditions. Our results highlight the need for improved robustness in CV-based AI systems to mitigate potential failures caused by such attacks, and the potential need for novel camera designs for critical security applications.

\section{RELATED WORKS}

Adversarial machine learning has uncovered a range of vulnerabilities in deep learning models, specifically in CV \cite{Kurakin2016AdversarialEI}. Early discoveries included small modifications to pixel values that could cause model output to be incorrect. These techniques, however, assume full access to the input pipeline (i.e. White Box) and are not generally applicable to physical-world settings.

New development of more realistic adversarial attacks that can be executed in real-world scenarios have begun to be heavily developed. Prior work has demonstrated that machine learning models for computer vision are vulnerable to physical-world adversarial attacks. These attacks typically rely on visual perturbations such as adversarial patches \cite{Brown2017AdversarialP}, camouflage patterns \cite{Zhang2018CAMOULP}, or light-based attacks \cite{Gnanasambandam2021OpticalAA} that mislead object detectors or classifiers in real-world scenes. Such attacks can require only partial (i.e. Grey Box) knowledge of the model, or no knowledge of the model (i.e. Black Box). 

In contrast to these purely visual attacks, \cite{Ji2021PoltergeistAA} introduced an acoustic adversarial attack that exploits inertial sensors within camera stabilization systems. By injecting acoustic signals at known resonant frequencies of the camera system, the attack induces false motion compensation, leading to motion blur and degraded object detection. Researchers showed a surprising high (nearly 100\%) attack success rate for causing popular object detection models to fail in classifying objects.

While impactful, \cite{Ji2021PoltergeistAA} and related studies focus on ultrasonic ($>$20 kHz) signals, which attenuate rapidly in air and thus only function at short range. In contrast, our work explores low-frequency acoustic signals that propagate over longer distances, hypothetically to affect a wider range of camera systems, and require less power to deploy.

The attacks depend on specifically crafted environments and their efficacy in uncontrolled environments remains under-explored. Our project introduces more realistic dynamic and noisy scenes which can account for more accurate situations in which attacks could be played out. As previously mentioned, we focus on lower frequencies due to their ability to propagate over longer distances  and more effectively traverse noisy environments.

We extend this line of inquiry by demonstrating that low-frequency acoustic resonance can be weaponized to attack the CV pipeline via mechanical coupling between sound waves and camera structures– even in the absence of image stabilization hardware. While previous work \cite{Ji2021PoltergeistAA} has shown that ultrasonic signals can disrupt stabilization sensors, to our knowledge, this is the first systematic exploration of low-frequency, long-range acoustic attacks on vision-based object detectors.

\section{PROPOSED METHOD}

\begin{figure}[H]
    \centering
    \includegraphics[width=0.8\columnwidth]{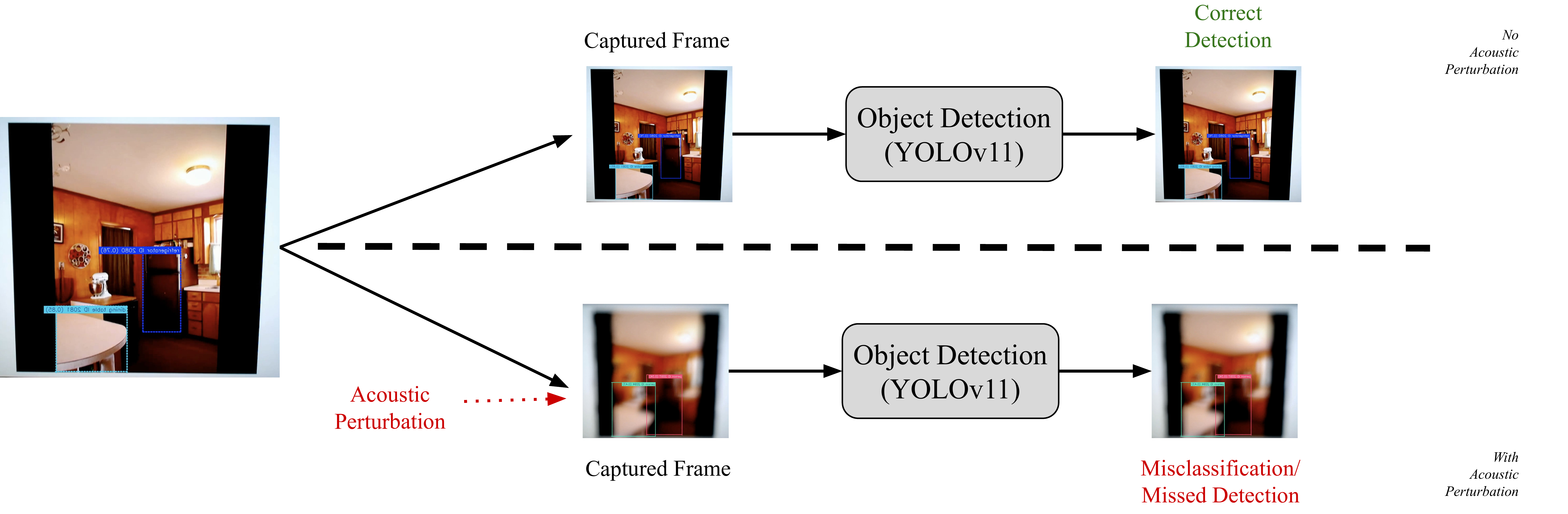}
    \vspace{0.3cm}
    \caption{Overview of the proposed acoustic attack pipeline. An acoustic signal induces mechanical vibrations in the camera, which distort the captured image during acquisition. These distortions propagate through the object detection model (YOLOv11) leading to incorrect predictions. The attack does not modify the digital input directly, but instead perturbs the physical sensing process, resulting in misclassification, missed detections, or spurious detections.}
    \label{fig:attack_pipeline}
\end{figure}

\subsection{Frequency Signal Generation}
\label{sec:signal_generation}

We generated acoustic waveforms using a Tektronix AFG3100 arbitrary function generator, which is capable of producing signals across a wide frequency range (from 5 Hz to 30 kHz). In our experiments, we focused on lower-frequency signals, as these are more likely to induce mechanical resonance and propagate effectively in real-world environments. Sine waves were selected due to their smooth waveform and energy concentration at a single frequency, which makes them ideal for studying resonant mechanical effects, similarly noted in prior injection studies \cite{Usb2014ArbitraryF} \cite{Roy2017BackDoorMM}. All signals were produced at a fixed amplitude of 8 Vpp and transmitted to a compact high-fidelity speaker, which served as the acoustic emitter.

To determine effective attack frequencies, we conducted preliminary sweeps by gradually increasing the output frequency in small increments (e.g., 10–50 Hz steps). These sweeps allowed us to identify frequency bands that induced visible vibrations or image distortions in the video stream. Identified “resonant bands” were prioritized for in-depth testing and analysis.

\subsection{Physical Equipment}
We used the \textbf{Logitech C930e Business Webcam} as the visual input device. While the C930e does not employ image stabilization, it still demonstrates sensitivity to physical vibration and mechanical resonance when subjected to external motion. This is likely due to the floating lens autofocus assembly, in which a lens is allowed to translate along the imaging axis to correct object focus. As the webcome is small, the autofocus assembly is inherently light and has little inertia, making it receptive to acoustic vibration. To simulate conditions under which stabilization systems might misinterpret physical disturbances, the webcam was rigidly mounted to the acoustic speaker. This configuration ensured that any induced motion would directly affect the camera’s frame, thereby allowing us to capture the physical effects of the sound waves on video output. The camera was positioned at a fixed distance from the display monitor throughout the entire experiment. 

\subsection{Real-Time Video Recording Under Acoustic Perturbation}
In order to replicate real-world object detection scenarios, we setup our web camera to record footage of a monitor which displayed various images from the COCO dataset \cite{Lin2014MicrosoftCC}. Each image was shown full-screen, and the webcam used captured a live feed of the screen. During the capturing, we emitted resonant band frequencies to test the effects of the sound pressure wave in real time. Each video was recorded for 3 seconds during which a frequency was continuously emitted. A total of 100 COCO images were tested, each under multiple frequency conditions (including baseline/ no-perturbation controls).

Each test condition consisted of:
\begin{itemize}
    \item Static COCO image displayed full-screen on the monitor.
    \item Acoustic signal played continuously for 3 seconds.
    \item Webcam recording live footage of the monitor during signal emission.
\end{itemize}

\subsection{Post-Processing for Object Detection}
We utilized the YOLOv11 objection detection model, which is a one-stage objection detection architecture that offers real-time detection. It performs detection as a regression task, directly predicting bounding boxes and class probabilities from full images in a single forward pass of the neural network \cite{Redmon2015YouOL}. 

The architecture includes three main components: the backbone, neck, and head. The backbone is what extracts important features from the input and is based on CSPNet. The neck is what aggregates features across multiple scales to detect objects varying in size. The head is what predicts the bounding boxes, scores, and probability\cite{Khanam2024YOLOv11AO} \cite{Zhang2025ZZYOLOv11AL}.

Once frequency inflicted videos were recorded, the following process was performed:
\begin{itemize}
  \item \textit{Cropping:} During recording, the camera recorded the entire monitor screen which included unwanted white space that did not relate to images. Therefore, images were cropped in order to show only the testing image.
  \item \textit{Inference:} YOLOv11 was ran on each video to calculate bounding boxes and confidence scores. 
  \item \textit{Pixel Extraction:} The average RGB and pixel area values of the entire video were extracted. Then the average RGB and pixel area values of each bounding box were extracted.
  \item \textit{Averages Calculations:} Aside from average RGB and pixel values, we found the averages of area, confidence, and accuracy. We summarized averages based on the grouping of each class within a tested frequency video.
  \item \textit{Master Dataset:} Once averages were found, a master dataset of each test frequency and test photo were combined and organized by both frequency and photo ID. 
\end{itemize}

The output of this process was used for the downstream analysis of image classifier quality and response to acoustic vibration.  

\section{EVALUATION METRICS}
\begin{figure}[H]
    \centering
    \includegraphics[width=0.7\columnwidth]{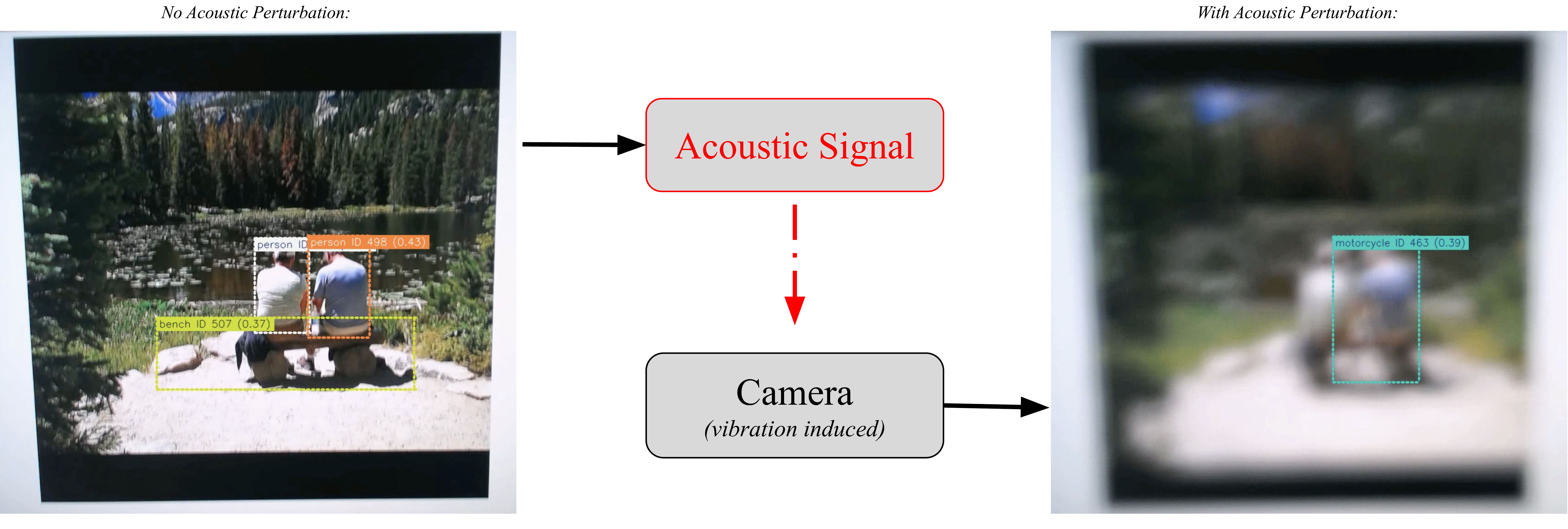}
    \vspace{0.3cm}
    \caption{Example of object detection degradation under acoustic perturbation. \textbf{Left}: baseline image captured without acoustic interference, where objects are correctly detected and localized. \textbf{Right}: image captured under acoustic excitation, where induced vibrations introduce blur and geometric distortion. These artifacts degrade the performance of the object detection model, resulting in incorrect classification and reduced detection rate.}    
    \label{fig:methodology}
\end{figure}

To evaluate the impact of acoustic perturbations on object detection performance, we conducted a series of controlled experiments across a range of low-frequency signals (0 Hz to 200 Hz). For each frequency, we recorded 3-second videos of COCO images displayed on a monitor, while the camera–mounted to a speaker was exposed to continuous sine waves. The YOLOv11 detector was then applied to each frame of the captured footage. Examples of the induced acoustic effect are shown in Figure \ref{fig:methodology}. Note the misclassification in the figure during acoustic irradiation. 

\subsection{Detection Stability}\label{ITH}
Initial sweeps showed that certain frequency bands caused significant drops in detection stability. In particular, frequencies between \textbf{20–30 Hz} and \textbf{155–180 Hz} produced noticeable visual distortion in the video frames, including frame jitter, blur, and geometric skew.

We decided to split the effects into three of the following categories:

\begin{itemize}

\item \textbf{Misclassification:} Objects were assigned incorrect labels compared to the ground truth. See Figure \ref{fig:missclassification}.

\begin{center}
\includegraphics[width=0.7\columnwidth]{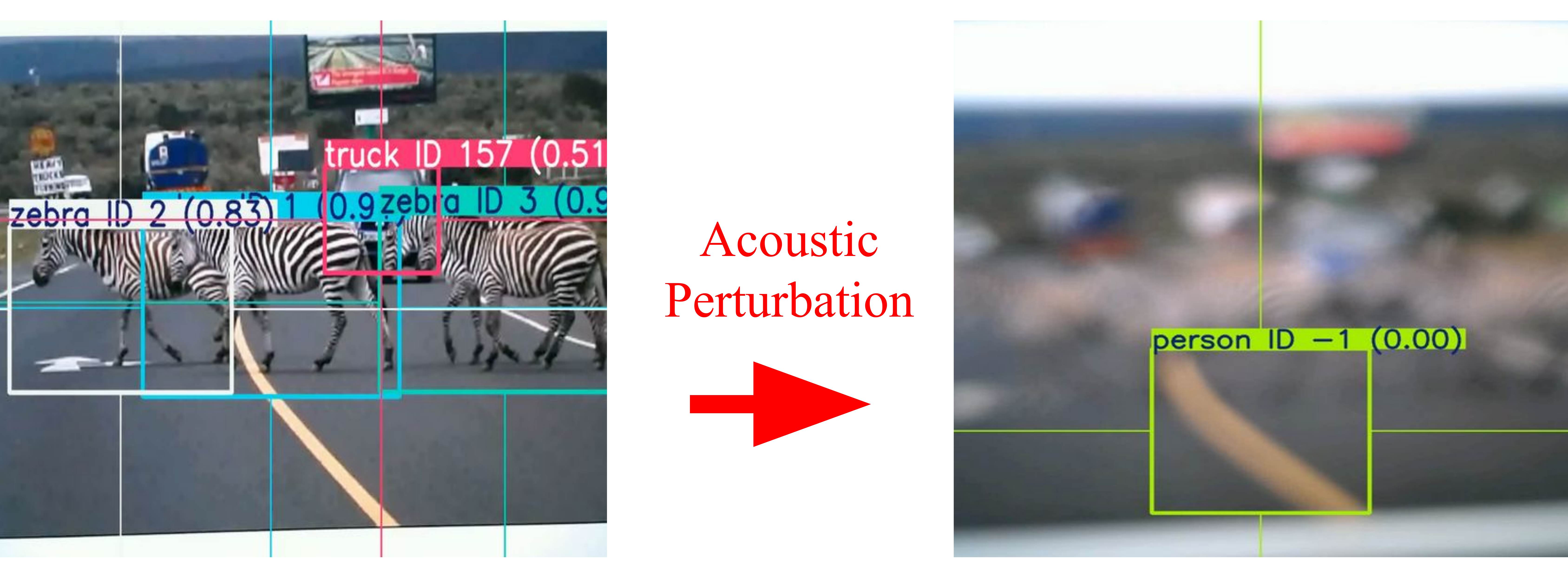}
\vspace{0.3cm}
\captionof{figure}{Misclassification: Originally correctly classified zebras now are no longer classified and instead, part of a road is incorrectly classified as a person.}
\label{fig:missclassification}
\end{center}

\begin{samepage}
\item \textbf{Suppression:} Legitimate objects were entirely undetected, with no bounding boxes or labels produced. See Figure \ref{fig:suppression}.

\begin{center}
\includegraphics[width=0.7\columnwidth]{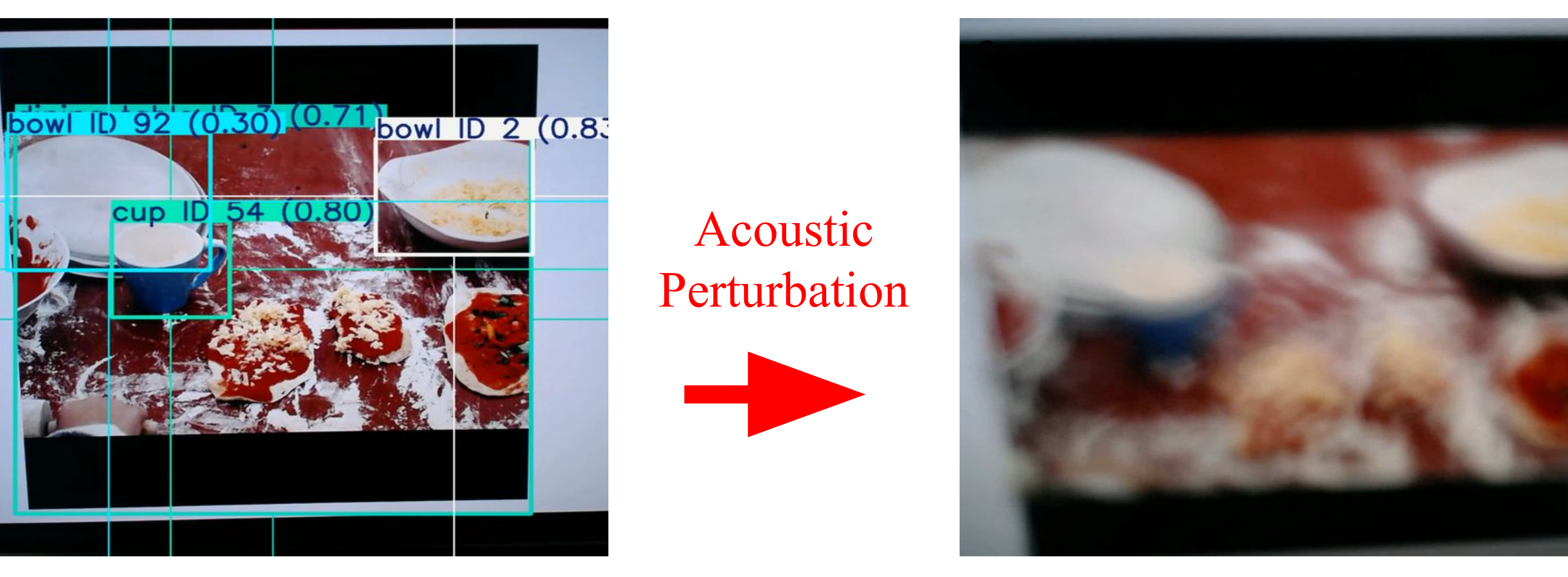}
\vspace{0.3cm}
\captionof{figure}{Suppression: Originally correctly classified bowls and cup are now no longer classified at all.}
\label{fig:suppression}
\end{center}
\end{samepage}

\item \textbf{Spurious Detection:} The model produced false positives, detecting objects where none were present. See Figure \ref{fig:spur_detect}.

\begin{center}
\includegraphics[width=0.7\columnwidth]{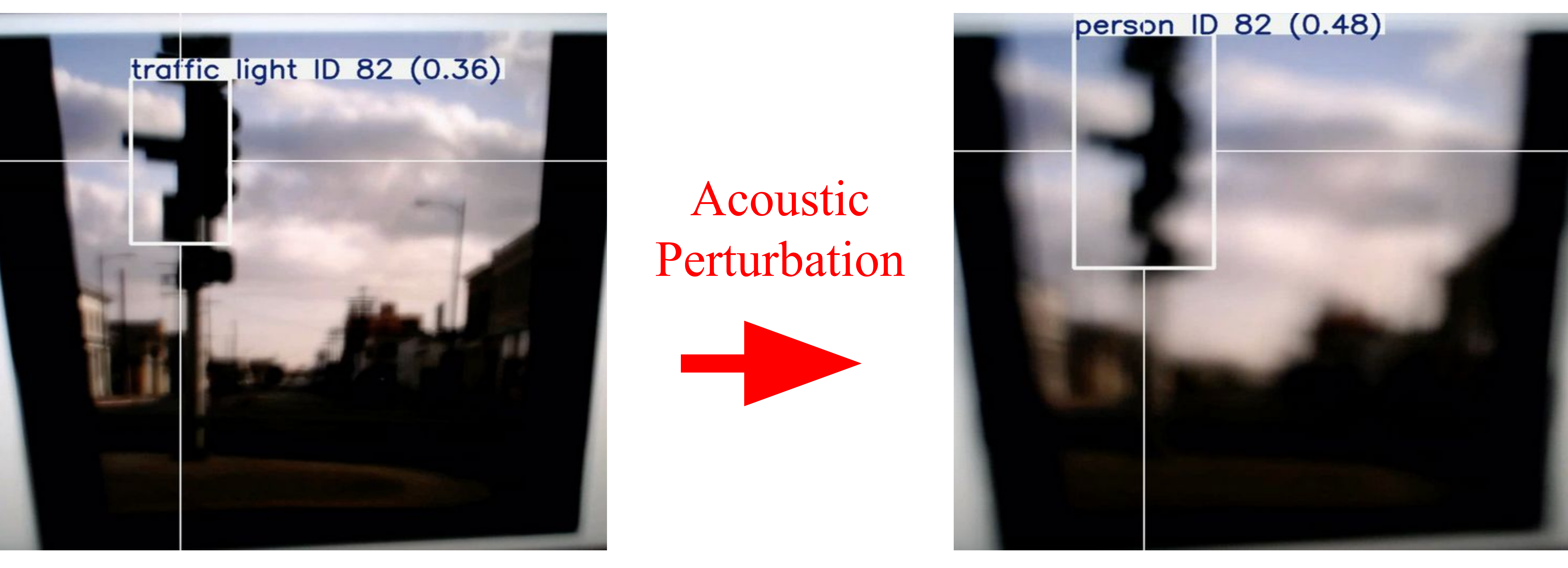}
\vspace{0.3cm}
\captionof{figure}{Spurious detection: Originally classified traffic light is now incorrectly classified as a person.}
\label{fig:spur_detect}
\end{center}

\end{itemize}

\section{ANALYSIS}
Our experimental results reveal that low-frequency acoustic perturbations can significantly impact the performance of state-of-the-art object detection models. We present our impact in terms of detection stability, model detection rate, error types, and image-level visual distortion under acoustic perturbation.

\subsection{Impact of Resonant Frequencies}

\begin{figure}[H]
    \centering
    \includegraphics[width=0.7\columnwidth]{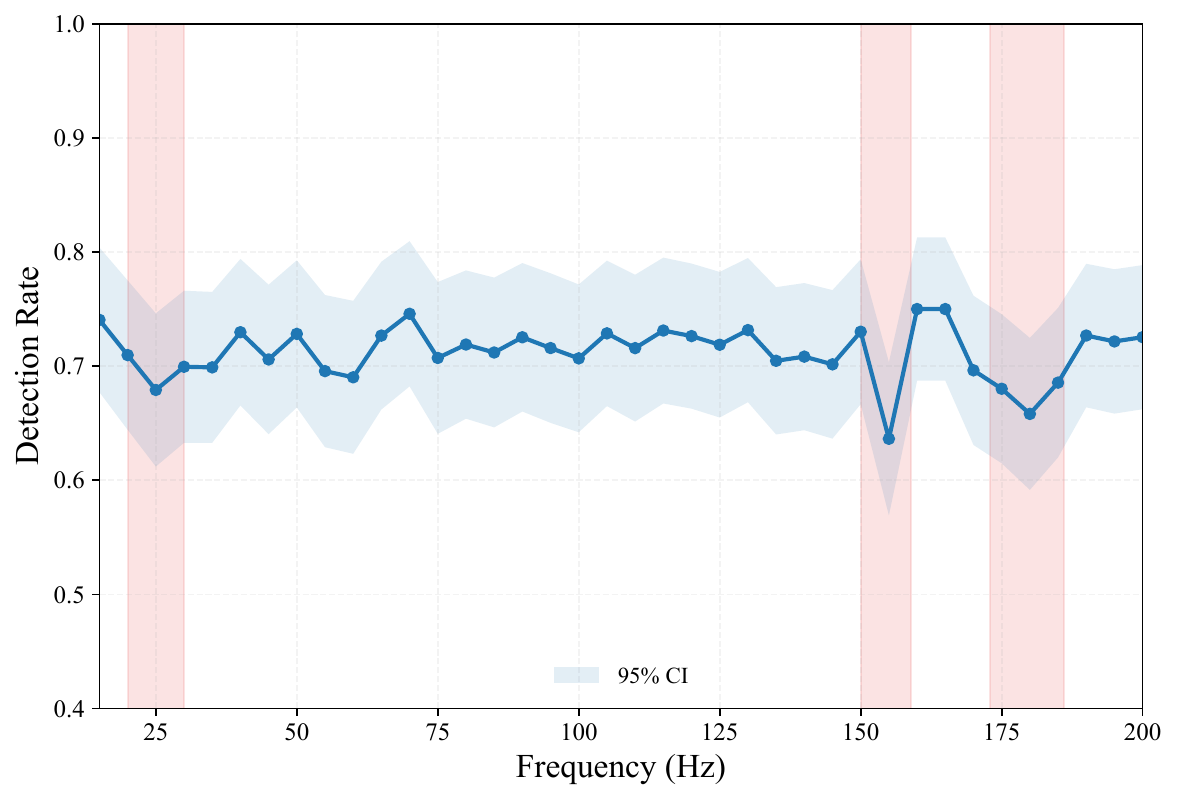} 
    \caption{Detection rate as a function of acoustic frequency. Detection rate is defined as the proportion of detections with confidence greater than or equal to 0.5. The blue shaded regions represent the 95\% confidence interval. Highlighted peach regions represent where acoustic perturbations produced the highest degradation in detection performance. Detection rate drops most within 20-30 Hz and 150-180 Hz, suggesting the presence of frequency-dependent vulnerabilities in the system.}
    \label{fig:detection_rate}
\end{figure}

Figure \ref{fig:detection_rate} shows detection rate as a function of the injected acoustic frequency. Two frequency bands—\textbf{20-30 Hz} and \textbf{155-180 Hz}—consistently resulted in a nearly 10\% drop in detection rate compared to baseline (no sound) conditions. These bands correspond to likely mechanical resonance points in the camera housing, which induced subtle vibrations visible in the captured frames. Similar resonance effects have been observed in prior sensor spoofing attacks \cite{Cao2021InvisibleFB} \cite{Trippel2017WALNUTWD}. These vibrations caused micro-movements in the large image sensor or lenses cause the motion blur, focal inconsistencies, or compression artifacts.

\subsection{Detection Errors and Visual Artifacts}

\begin{figure}[H]
    \centering
    \includegraphics[width=0.6\columnwidth]{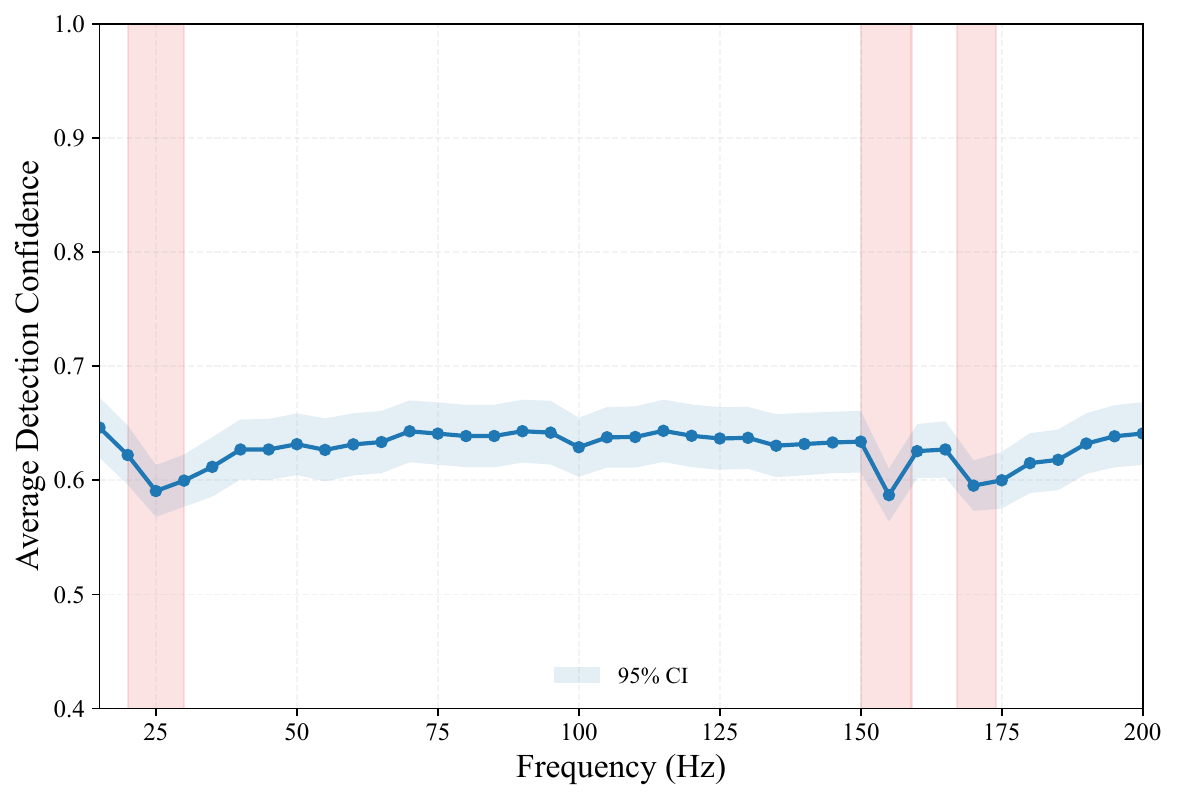} 
    \caption{Average detection confidence as a function of acoustic frequency. The blue shaded regions represent the 95\% confidence interval. Highlighted peach regions represent frequency ranges associated with increased distortion. While confidence remains relatively stable across most frequencies, there are noticeable drops around 20-30 Hz and 150-180 Hz, similar to our detection rate shown in Figure \ref{fig:detection_rate}. These drops indicate that acoustic perturbations degrade our model's certainty alongside detection failures.}
    \label{fig:avg_confidence}
\end{figure}

As shown in Figure \ref{fig:avg_confidence}, acoustic perturbation not only reduces detection rates but also significantly lowers model confidence scores. At resonant frequencies, the average confidence dropped by over 7\%, indicating higher model uncertainty. Both metrics exhibit noticeable degradation within specific frequency bands. These degradation suggests that acoustic perturbations affect the visual pipeline in a frequency-dependent manner.

Misclassifications and phantom detections suggest that spatial noise and distortion are enough to confuse YOLOv11’s feature extraction layers. This supports the hypothesis that early convolutional layers, which are highly sensitive to texture and shape cues, can be disrupted by even minor image corruption \cite{Szegedy2013IntriguingPO}. Notably, the hallucination of objects in blank areas may stem from pattern-like distortion introduced by motion artifacts, which the network erroneously interprets as valid object features.

\subsection{Robustness and Model Sensitivity}

Our findings raise important concerns about the robustness of real-time CV systems in physical-world environments. Even without direct access to the model or its parameters, external actors may be able to interfere with its perception pipeline using basic audio equipment. These attacks are stealthy (potentially inaudible outside a target region if multiple emitters are used) and do not rely on adversarial perturbations to the input image, making them difficult to detect or mitigate using traditional AI defenses.

\section{Discussion}
The primary vulnerability behind this attack is the acoustic excitation of mechanical resonances in the camera system. Lens assembly, CMOS sensor mount, and stabilization components can all vibrate in response to low-frequency sound waves. These micro-vibrations introduce subpixel motion and spatial distortion, which propagate through the CV pipeline. 

Our findings suggest that commodity of cameras are vulnerable due to constitutions, lack of damping, and proximity to mechanical resonant frequencies in the audio spectrum. Similar findings have been reported in physical-domain adversarial attacks, where sound waves alter camera input before digitization \cite{Cao2021InvisibleFB} \cite{Trippel2017WALNUTWD}.

YOLOv11, like many modern CNN-based detectors, is sensitive to small changes in edge clarity, texture gradients, and object boundaries. The visual artifacts introduced by acoustic vibrations can significantly distort these features. Notably, early convolutional layers—responsible for detecting edges and shapes—are most impacted, which leads to errors that persist throughout the network. However, because the attack does not manipulate the input image pixels directly, but rather distorts the physical capture process, traditional adversarial defenses (e.g., adversarial training, input pre-processing) are ineffective \cite{Carlini2017AdversarialEA}.

This simple attack imposes a significant threat to real-world AI systems. Unlike digital adversarial examples which may require extensive knowledge of pipelines and models, acoustic perturbations can be delivered without direct access to the vision system of software and without the requirements of extensive setups or tools. They also require new defenses, which are not currently an active area of research. 

\section{Conclusion and Future Work}
We have presented a novel class of acoustic attacks on computer vision systems that operate in the low-frequency domain. Unlike prior work based on ultrasonic interference, our approach exploits mechanical resonances in commodity cameras using audible or near-audible signals that can travel significantly further in air. These signals cause physical vibrations that introduce motion blur, spatial jitter, and frame distortion—ultimately degrading the performance of object detection models such as YOLOv11.

Our physical experiments show that even cameras without stabilization hardware are susceptible to this form of interference, suggesting broader applicability. The attack is hardware-agnostic, stealthy, and requires no access to the internal CV model or image input.

While our setup effectively demonstrates the vulnerability of YOLOv11 to acoustic interference, there are limitations. We focus only on a single camera model (C930e), and one that does not have a fully dedicated image stabilizer. In future runs, we hope to compare several camera models to capture what our attacks would look like in different scenarios, especially those that use more advanced stabilizers. Our second limitation was the fact that the acoustic source was fixed and directional; real-world interference may come from multiple or mobile sources.

By demonstrating that vulnerabilities exist at the physics-sensor interaction stage of image classifiers in the AI perception stack, we hope to contribute to a broader understanding of the physical security risks facing modern computer vision systems.

\acknowledgments 

This manuscript has been assigned LA-UR-26-22814. Research presented in this paper was supported by the Information Science and Technology Institute's (ISTI) Cybersecurity Science Research Program (CSRP). The authors also acknowledge Chris Rawlings for facilitating conference support. LANL is operated by Triad National Security, LLC, for the National Nuclear Security Administration of the U.S. Department of Energy (Contract No. 89233218CNA000001).
\nocite{*}
\bibliography{report} 
\bibliographystyle{spiebib} 

\end{document}